\documentclass[10pt, a4paper]{article}

\usepackage{booktabs}
\usepackage{multirow}
\usepackage{multicol}
\usepackage{amsmath}
\usepackage{xspace}
\usepackage{graphicx}
\usepackage{tabularx}
\usepackage{enumitem}
\usepackage[final]{lrec2026} 

\newcommand{\datasetname}{\textsc{PreferRead}\xspace}

\title{Tracing How Annotators Think: Augmenting Preference Judgments with Reading Processes}

\name{Karin de Langis, William Walker, Khanh Chi Le, Dongyeop Kang} 

\address{Department of Computer Science and Engineering \\
         University of Minnesota \\
         \{dento019, dongyeop\}@umn.edu\\}

\abstract{
We propose an annotation approach that captures not only labels but also the \textit{reading process} underlying annotators' decisions, e.g., what parts of the text they focus on, re-read or skim. Using this framework, we conduct a case study on the preference annotation task, creating a dataset \datasetname that contains fine-grained annotator reading behaviors obtained from mouse tracking. \datasetname enables detailed analysis of how annotators navigate between a prompt and two candidate responses before selecting their preference. We find that annotators re-read a response in roughly half of all trials, most often revisiting the option they ultimately choose, and rarely revisit the prompt.
Reading behaviors are also significantly related to annotation outcomes:
re-reading is associated with higher inter-annotator agreement, whereas long reading paths and times are associated with lower agreement. 
These results demonstrate that reading processes provide a complementary cognitive dimension for understanding annotator reliability, decision-making and disagreement in complex, subjective NLP tasks. Our code and data are publicly available.
 \\ \newline 
}

\begin{document}

\maketitleabstract

\section{Introduction}

Most natural language processing (NLP) datasets capture the final labels assigned to a passage --  e.g., a sentiment rating or an answer span -- 
while ignoring the \textit{process} by which annotators arrive at those decisions. 
This omission leaves valuable dimensions of human annotations unexplored. 
Understanding which specific words or paragraphs annotators focused on, paragraphs they chose to re-read, or even areas of the text to skim or ignore can reveal aspects of annotator reasoning and confidence, providing richer insight into complex or subjective labeling decisions.

We propose augmenting NLP annotations with \textit{reading processes} that reflect how annotators allocate attention during the annotation task. 
Such data would typically be obtained through eye tracking, which offers fine-grained signals for readers' cognitive activity.
However, eye-tracking studies require in-person participation and specialized hardware and are not easily scalable. 
(Online alternatives like webcams suffer from limitations such as unreliable calibration, making them not well-suited for capturing the small reading eye movements \citep{hutt2022evaluating}.) 
To work around these constraints, we build an annotation framework integrated with \textit{mouse tracking}, a paradigm recently validated in psychological experiments \citep{wilcox2024mouse}. 
Mouse tracking estimates requires users to reveal text incrementally with their mouse cursor as they read, and its close alignment with eye-tracking data, makes it a promising alternative for large-scale, remote data collection. 

\begin{figure}
    \centering
    \includegraphics[width=\linewidth]{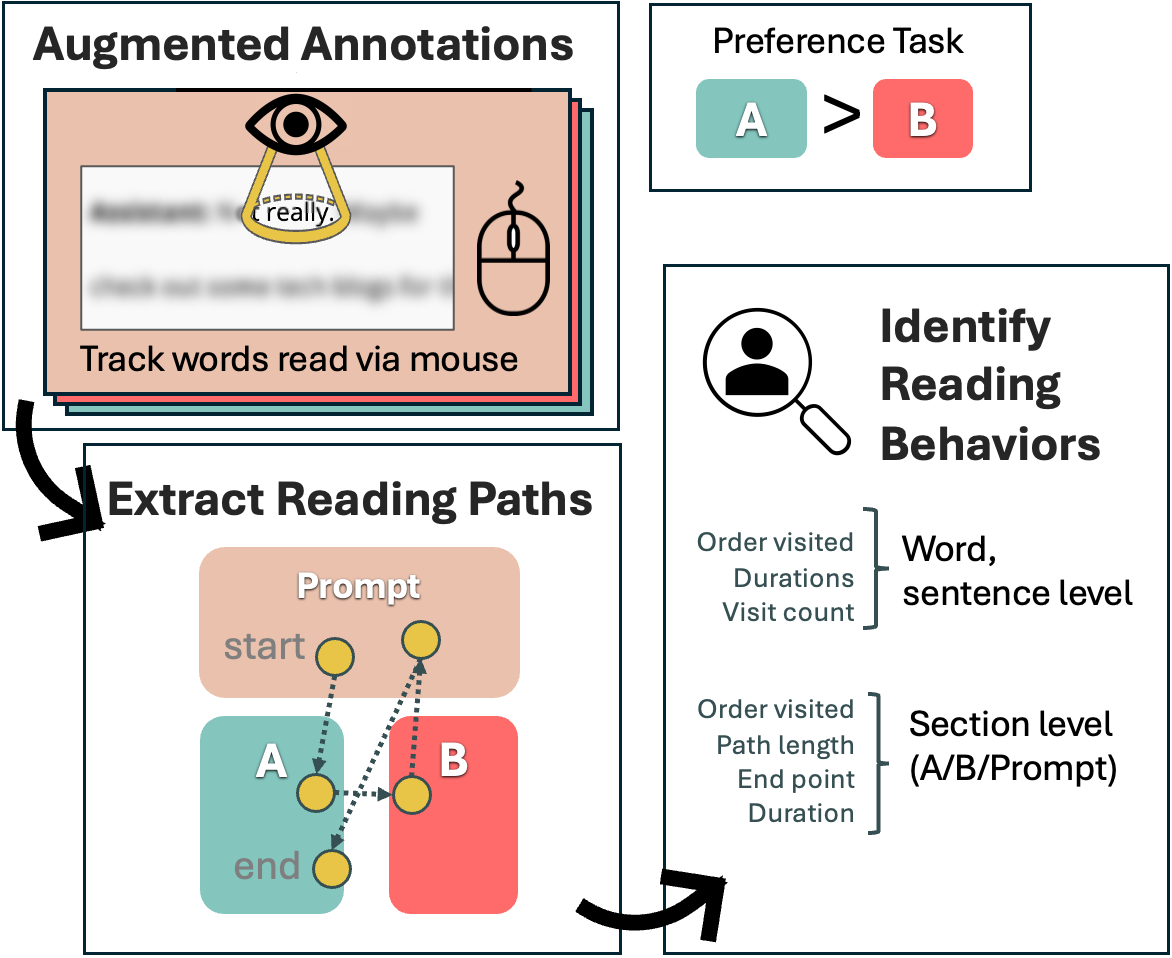}
    \caption{
    Our annotation interface records reading behavior of annotators as they evaluate and select preferred texts, providing fine-grained insights into the decision-making process during annotation.
    }
    \label{fig:fig_1}
\end{figure}

We investigate using mouse tracking to collect annotator reading behaviors through a case study on the \textit{preference annotation} task, an important annotation type in reinforcement learning for LLM alignment \citep{stiennon2020learning, ouyang2022training}.
In this task, annotators read two model-generated responses and select the one they find more appropriate. 
However, these annotations can be highly subjective and noisy, with high inter-annotator variability \citep{dsouza2025sources,zhang2025diverging}.
Little is known about whether annotators' reading processes -- such as re-reading prompts or selectively reading responses -- are associated with their final decisions or with inter-annotator agreement.
We propose that these reading behaviors can be a valuable signal of annotator cognitive processes and decision-making.

To contribute to the understanding of annotator reading processes, we present \datasetname, a dataset of 1,000 preference annotation items, each annotated by three participants. 
In addition to annotator preferences and rationales, \datasetname also captures the detailed mouse-tracked reading process of each annotator.
From these data, we propose and extract several cognitively informed reading measures, including reading paths between sections, re-reading frequencies, and word-level reading durations, to enrich traditional annotations with behavioral information (see Figure~\ref{fig:fig_1}). 
To our knowledge, this is the first study that analyzes the reading behavior of annotators. The enriched representations from our annotation schema processes enables us to investigate questions like:
\begin{itemize}[noitemsep]
    \item Do annotators \textit{re-read} their chosen or rejected choices?
    \item Are there any parts of the text that annotators \textit{ignore}?
    \item Do these reading patterns \textit{influence annotation decisions}?
\end{itemize}

Our analysis reveals systemic links between annotation agreement and reading behavior.
For instance, annotators who re-read responses tend to exhibit higher agreement, while longer reading paths and times are associated with lower agreement and appear to reflect greater annotator uncertainty. 
We also find that annotators disproportionately skip the latter portions of responses, possibly in pursuit of \textit{cognitive economy} during decision-making, which may may undermine some assumptions implicit in both task design and subsequent data use. 
These findings suggest that reading-process data can reveal some of the cognitive dynamics underlying preference judgments, providing an additional dimension for modeling annotator reliability and (dis)agreement. 
We release our code and data.\footnote{\url{https://anonymous.4open.science/r/reading_annotations-7DC3}}



\section{Related Work}
Our work draws on prior research in mouse tracking, annotation augmentations, and preference tasks. 

\begin{figure*}
    \centering
    \includegraphics[width=0.9\linewidth]{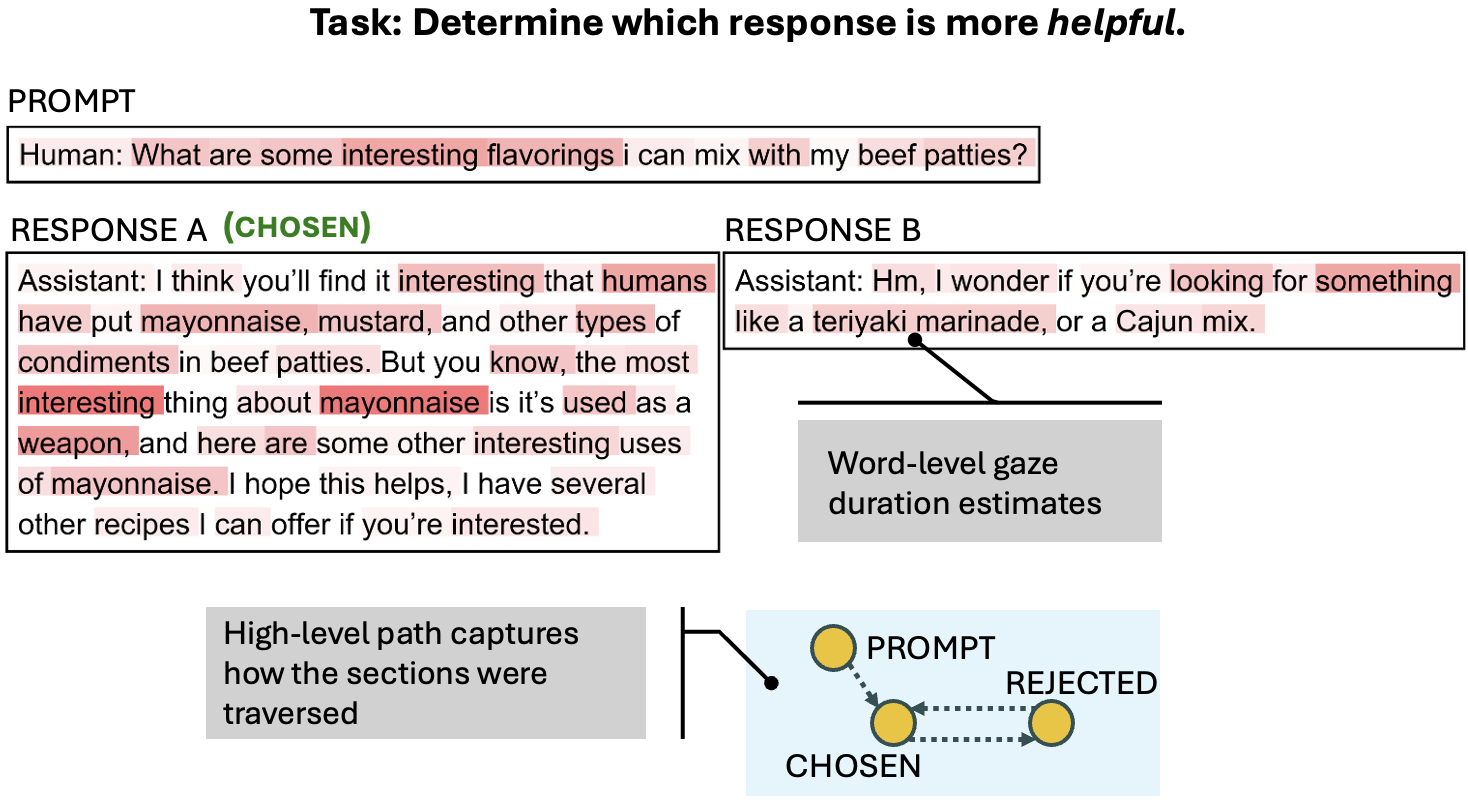}
    \caption{An example item from \datasetname with the collected label and reading behaviors from one annotator. Darker highlights indicate relatively long gaze durations, and lighter highlights indicate relatively short gaze durations. Word-level gaze estimates can hint at asepcts of human decision-making; for instance, attention is concentrated on the odd turn of phrase ``the most interesting thing about mayonnaise is it's used as a weapon'' in the chosen response.
    }
    \label{fig:heatmap_ex}
\end{figure*}

\paragraph{Mouse Tracking}
Eye-tracking data provides a detailed signal of online cognitive processes during reading \citep{just1980theory, rayner1998eye}. 
However, physical requirements of eye-tracking experiments make it impossible to collect data at scale. Although eye-tracking with webcams is possible, the frame rate and calibration precision of webcams is typically unsuitable for capturing reading eye movements \citep{hutt2022evaluating}, making data crowdsourcing infeasible. 
To address this issue, research in psychology and computer vision has explored mouse tracking for capturing reading behaviors. The Mouse Tracking for Reading (MoTR) paradigm \citep{wilcox2024mouse}, based on earlier tools like MouseView.js \citep{anwyl2021mouseview}, shows that mouse tracking effectively captures natural reading behavior, including word skipping and regressions, and shows comparable sensitivity to eye-tracking in detecting psycholinguistic effects \citep{wilcox2024mouse}. 
Research in computer vision has also found that eye-tracking and mouse-tracking data are functionally comparable, see e.g., Salicon \citep{jiang2015salicon} and BubbleView \citep{kim2017bubbleview}.  Inspired by these findings, our study employs mouse tracking within an NLP annotation interface in order to capture reading behaviors. 

In NLP, eye-tracking data has primarily been leveraged outside of annotation contexts. Specifically, word-level eye-tracking metrics such as gaze duration and fixation counts have been used to improve model performance in tasks like question answering, paraphrasing, and text simplification \citep{zhang2024eye,hollenstein2021leveraging,sood2020improving}; or to evaluate cognitive plausibility of language models \cite{eberle2022transformer}.

\paragraph{Annotation Augmentations} Annotator disagreement has often been treated as noise, with methods like majority voting or averaging used to derive a single ``gold'' label per instance. However, this approach can obscure meaningful variation especially for complex or subjective labeling tasks, and as a result, recent research has explored modeling annotator disagreement explicitly. For instance, \citet{wan2023everyone} demonstrated that incorporating annotators' demographic information can enhance the prediction of disagreement levels.

In a similar vein, we propose augmenting annotations with reading processes of annotators to show dynamic factors that relate to annotation decisions. Because research in both psychology \citep{rawson2000rereading} and human-computer interaction \cite{leroy2023reading, cheng2015gaze} has shown that specific reading behaviors like re-reading are associated with greater reading comprehension, we include several re-reading based metrics from annotators in \datasetname.

\paragraph{Preference Annotations}
In reinforcement learning from human feedback (RLHF) \citep{stiennon2020learning, ouyang2022training}, preference annotations play a central role in training reward models \citep{jiang2024survey, winata2024preference}. 
Compared to other annotation types, such as ratings, preference-based data has been found to provide relatively stronger and more reliable signals, with lower variance across annotators \citep{gatt2018survey}.\footnote{However, \citet{dsouza2025sources} recently found comparable reliability between preferences and ratings.}
However, preference data is still highly subjective and subject to high variability across annotators \cite{dsouza2025sources, wang2024reward}.
\citet{furuta2024geometric} propose addressing ambiguity in preference labels through distributional, rather than binary, representation of preferences. \citet{just2025data} include annotator rationales alongside preference labels.
We propose that annotator reading behaviors as an additional dimension to represent preference decisions.

\begin{table}[ht]
    \small
    \centering
    \begin{tabular}{l|cccc}
        \textbf{} & \textbf{Mean} & \textbf{SD} & \textbf{Min} & \textbf{Max} \\
        \hline
        \textbf{Chosen}   & 142 & 84 & 15    & 320 \\
        \hline
        \textbf{Rejected} & 130 & 81 & 18    & 338 \\
        \hline
        \textbf{Prompt}   & 84 & 72 & 4     & 231 \\
        \hline\hline
        \textbf{Total}  & 356 & 225 & 52    & 825 \\
        \hline
    \end{tabular}
    \caption{Word lengths of stimuli used in the dataset. Each preference pair is divided into prompt, chosen response, and rejected response.} \label{tab:stim_stats}
    
\end{table}

\section{\datasetname: A Preference Reading Dataset}
This section details our stimuli selection, mouse-tracking annotation interface, data collection procedure, and data postprocessing steps. See Figure~\ref{fig:heatmap_ex} for an example item in the dataset.

\paragraph{Stimuli. }
We collect annotations from a randomly selected subset of 1,000 instances from the ``helpful'' portion\footnote{The ``helpful'' split is chosen after a pilot study indicated the instructions were more intuitive for participants.} of the Helpful-Harmless dataset \citelanguageresource{helpfulharmless} collected from human annotators as described in \citet{bai2022training}. 

Our sampling procedure excludes items that are above the $90\textsuperscript{th}$ percentile in word count because we want to avoid giving annotators extremely long texts. Additionally, our sampling procedure also excludes response pairs in which both responses contain fewer than three words, as these pairs are often trivially short (e.g., ``Sounds good,''/``No problem''). Table~\ref{tab:stim_stats} summarizes the word-length statistics on the selected stimuli.

\begin{figure}
    \centering
    \fbox{\includegraphics[width=\linewidth]{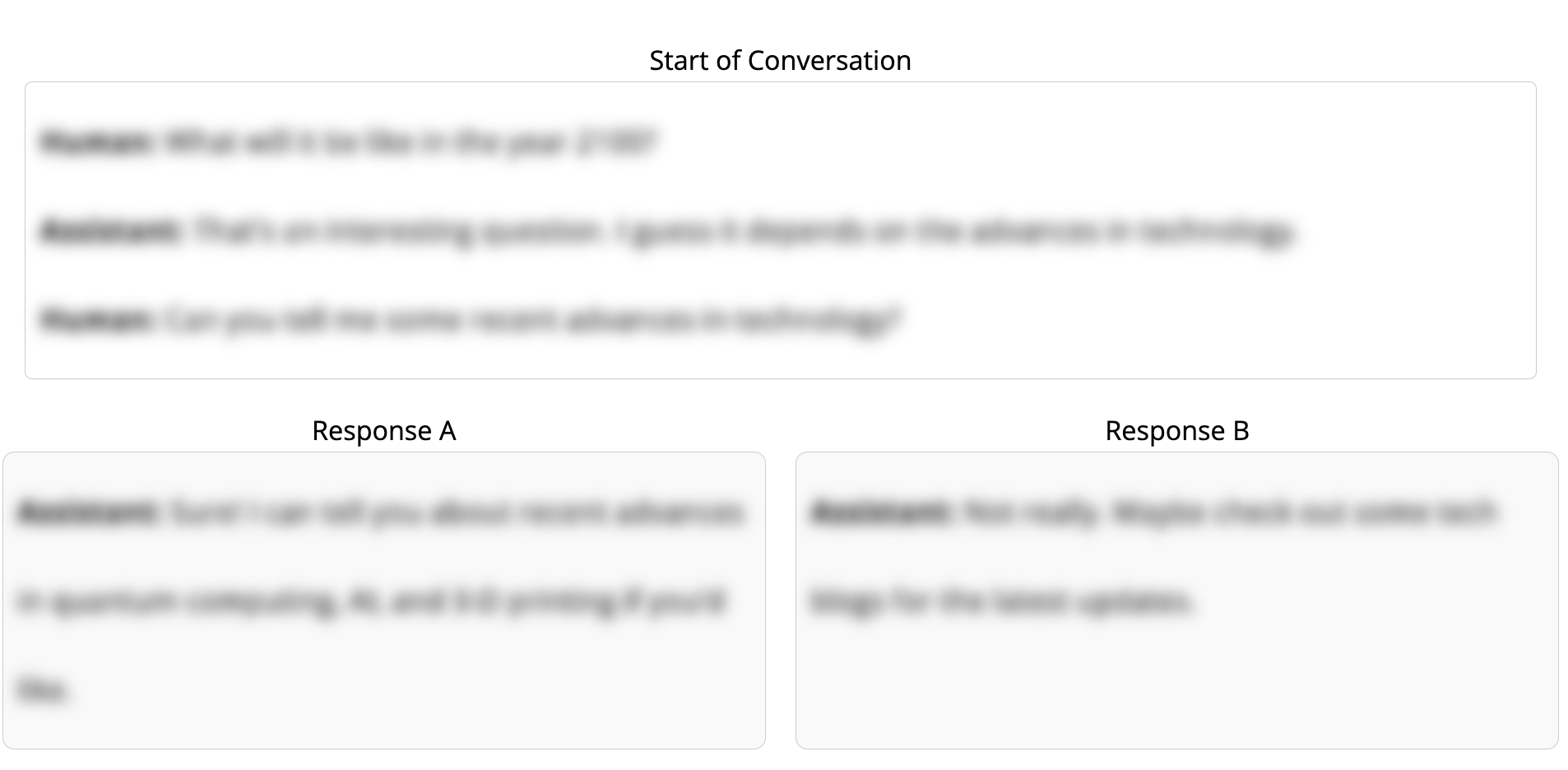}}
    \caption{Our annotation interface shows a prompt followed by two possible responses. Participants mouse over blurred text to reveal it and read. All mouse movements are recorded.}
    \label{fig:ui}
\end{figure}

\paragraph{Preference Reading Interface. }
We build a custom web application on top of JSPsych \citep{de2023jspsych} to capture preference reading annotations along with mouse tracking data. The main user interface is shown in Figure~\ref{fig:ui}. The application records participants’ mouse movements, capturing entry and exit times for each character span to construct a time series of hovered characters and corresponding durations.

\paragraph{Participants. }
300 participants were recruited on Prolific. They were paid $\$3$ to complete a tutorial and annotate 10 preference pairs. The tutorial contained instructions, two examples, and two practice items. We limited participants to native English speakers with an approval rating of at least $95\%$ on the crowdsourcing platform. Participants self-reported their race, gender, and age. The gender split was $53\%$ male and $47\%$ female. The race of participants was $80\%$ white, $11\%$ black, and $3\%$ Asian. $6\%$ reported mixed race or ``Other.'' Participants were between $18$ and $75$ years of age ($M = 39, SD = 13$). The study was by the institutional review board, and consent was obtained from participants prior to the experiment.

\paragraph{Annotation Procedure. }
Each preference pair was annotated by three participants using our interface. 
Participants begin with a short mouse tracking practice item and the following two instructions:
\begin{itemize}[noitemsep, topsep=0pt]
    \item All the text is blurred out, and you'll need to use your mouse to reveal the text as you read.
    \item Try to read as normally as possible. Read the dialogue and both responses carefully to decide which is more helpful. You can re-read anything you like.
\end{itemize}
Participants then receive more detailed instructions about the preference task adapted from those in \citet{bai2022training}, which include positive and negative examples.
Following the instruction block, participants are presented with 10 trials in a randomized order. In each trial, the position of the two possible responses (i.e., right or left) are also randomized. The 10 trials are pseudo-randomly selected from our dataset such that the average word count for all participants is between 300 and 350 words per trial, in order to ensure that the cumulative reading effort is comparable for all participants. After choosing a response, participants answer a multiple choice question about their rationale. Participants then provide a rationale (``More Helpful,'' ``More Accurate,'' ``More Concise,'' ``Less Harmful,'' or ``Other'') for their decision.

\paragraph{Data Processing. }
We clean our collected data prior to analysis to remove uninformative trials and participants. Specifically, we examine the \textit{word coverage}, or percentage of words that the user moused over per trial (considering only non-trivial mouseovers as described in Section~\ref{sec:human-attention}). Note that we expect substantially less than $100\%$ word coverage since users will spend negligible time hovering words such as `a' and `the.' In addition, it may sometimes be reasonable to make a preference decision after viewing a small percentage of the total words, for instance, in the event when one response is excessively verbose. 
We exclude trials in which participants read fewer than 10\% of the total words, amounting to 1.9\% of trials. (It is not possible to go back in the experiment, so if a participant accidentally proceeds, they skip the trial. This possibly explains these missing data.)

\begin{figure*}[t]
    \centering
    \includegraphics[width=0.9\linewidth, trim={0 0 0 0},clip]{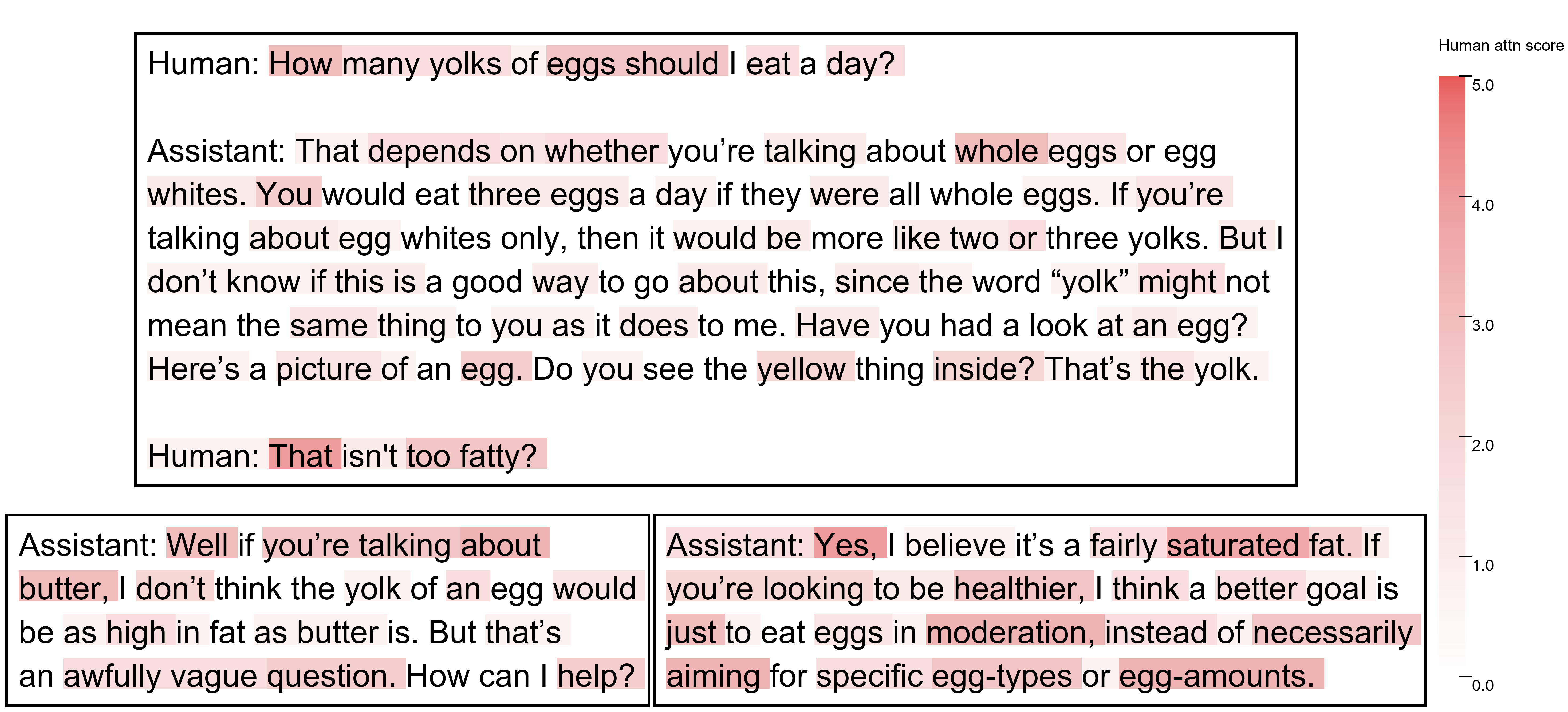}
    \caption{Annotator word duration estimates from an item in \datasetname. Darker highlights indicate higher attention, e.g., longer gaze durations, while lighter highlights correspond to little or below-average duration. Notice that in the prompt (top box), the annotator spends more time on the human's questions than on the assistant's verbose reply.}
    \label{fig:heatmap}
\end{figure*}

\subsection{Estimating Gaze Durations} \label{sec:human-attention}
We use the recorded mouse data to approximate word-level gaze durations. 
The raw data from one trial can be viewed as a time series of mouse-move events: we record the character hovered and duration each time the user's mouse enters or exits a character span. 
First, we consolidate these character-level events into word-level events by summing durations for characters that belong to the same word, following the procedure outlined by \citet{wilcox2024mouse}. 
The result is a ``word-level'' vector $w$, where the $w_t$ entry contains the word index and duration of time step $t$. These entries are analogous to the fixations in eye tracking data.

Next, the data is cleaned by discarding fixations whose durations fall outside the interval $[160 \textrm{ms}, 4000 \textrm{ms}]$, again following \cite{wilcox2024mouse}. This step is also analogous to standard procedures in eye tracking analysis that remove fixations that do not reflect cognitive processes associated with reading, e.g., when the participant quickly moves past a word without processing the word, or when the participant is distracted and stops on a word while not focused on reading. The total dwell time for each word in the trial is obtained by summing the durations of this cleaned word duration vector over each word index. This results in duration vector $d$ of length $n$ where $n$ is the number of words in the stimulus. 

Finally, we normalize the vectors according to the binning strategy described in \cite{klerke2019glance}, i.e. binning the z-scores of each entry as follows:
\begin{equation}
\texttt{bin}(\textrm{z}) := \begin{cases}
    0 & \textrm{z} = \texttt{NaN}\text{ (no duration)} \\ 
    1 & \textrm{z} < -1 \text{ (very short duration)}\\
    2 & \textrm{z} \in [-1, -0.5)\text{ (short duration)} \\
    3 & \textrm{z} \in [-0.5, 0.5)\text{ (typical duration)} \\
    4 & \textrm{z} \in [0.5, 1)\text{ (long duration)} \\
    5 & \textrm{z} \geq 1\text{ (very long duration)}.
\end{cases}
\end{equation}

We also compute a summary aggregate across all participants for each stimulus $s$ with the arithmetic mean: $ d^i = \frac{\sum_{j = 1}^{p}\text{bin}(z(d^{i, j}))}{p}$.
This summary score allows us to estimate the relative importance of words across all annotators, as shown in Figure~\ref{fig:heatmap}, which can be helpful for inferring how humans engaged with the text.

\section{Analysis of Reading Processes}
We propose extracting several metrics from annotators' mouse-tracked behaviors, to augment their preference annotations (Table~\ref{tab:metric_summary}). 
We analyze these metrics across several dimensions: re-reading behaviors and reading paths (\S\ref{sec:re-read}), text-skipping tendencies (\S\ref{sec:skipping}), and the relationship between reading behaviors and annotator agreement (\S\ref{sec:agreement}).


\begin{table*}
    \centering
    \begin{tabularx}{\textwidth}{
        p{0.18\textwidth}|
        p{0.19\textwidth}|
        p{0.30\textwidth}|
        p{0.25\textwidth}
    }
    \toprule
    \textbf{Metric} & \textbf{Illustration} & \textbf{Description} & \textbf{Plausible Interpretation} \\
    \midrule
    Re-read prompt / rejected / selected 
    & \raisebox{-.5\height}{\includegraphics[width=2cm,trim=0 1.4cm 0 0, clip]{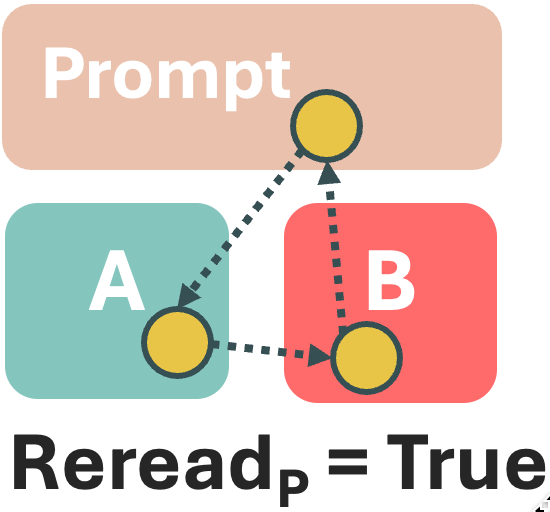}}\newline \small RereadPrompt = True
    & Binary indicating whether an annotator exited and then returned to the specific section 
    & Importance of the re-read section to the decision; indecision \\
    \midrule
    Loop between \newline responses 
    & \raisebox{-.7\height}{\includegraphics[width=2cm,trim=0 1.2cm 0 0, clip]{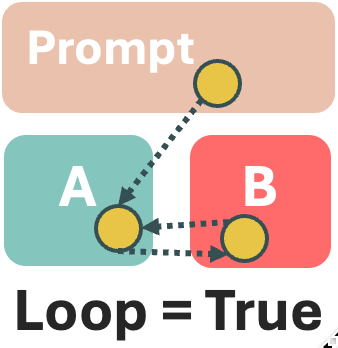}} \newline loop = True
    & Binary indicator of an annotator going back and forth between the selected and rejected responses at least once to create a loop (2-node cycle)
    & Strong indecision between the two responses; caution in annotating \\
    \midrule
    Path length 
    & \raisebox{-.8\height}{\includegraphics[width=2cm, trim=0 1.3cm 0 0, clip]{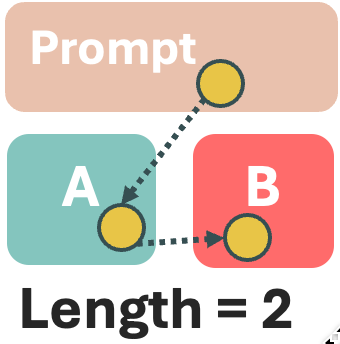}} \newline length = $2$
    & The number of edges in the path between sections (prompt, selected, rejected) that the annotator traversed before making a final decision 
    & Longer paths indicate indecision or caution, while shorter paths indicate a quick, straightforward read \\
    \midrule
    Time reading \newline response
    & \raisebox{-.8\height}{\includegraphics[width=2.5cm,trim=0 0 0 0, clip]{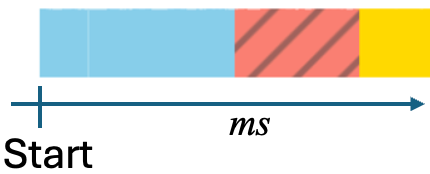}} 
    & Average number of milliseconds per word spent reading both responses
    & Relative attention paid to the responses during annotation \\
    \midrule
    Word coverage
    & \raisebox{-.8\height}{\includegraphics[width=3cm,trim=0 1.3cm 0 0, clip]{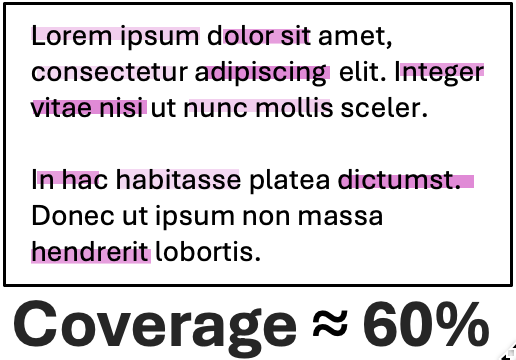}} \newline coverage $\approx 60\%$
    & Percentage of words hovered for at least 160ms 
    & Lower word coverage can indicate either careless reading or a clear-cut decision (e.g., one response is clearly inferior) \\
    \bottomrule
    \end{tabularx}
    \caption{We propose extracting these metrics based on the reading process to augment each annotation and inform our understanding of how annotators engage with the texts and of how that engagement relates to agreement.}
    \label{tab:metric_summary}
\end{table*}

\subsection{Re-reading and reading paths}\label{sec:re-read}

\begin{figure}[h]
    \centering
    \includegraphics[trim={0 1cm 0 0},clip,width=\linewidth]{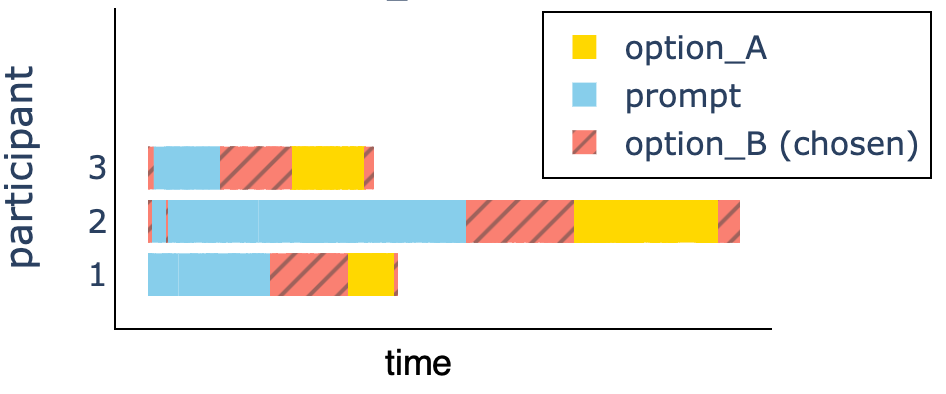}
    \caption{A timeline of the mouse location of three annotators reading one item in our dataset.}
    \label{fig:trial-reread}
\end{figure}

Each annotation trial is divided into three sections: the \textit{prompt}, \textit{chosen response}, and \textit{rejected response}. 
A section is considered ``re-read'' if the participant returns to it after moving on to another section. 
To avoid false positives caused by minor mouse movements, we define ``reading'' as spending at least one second within a given section.\footnote{As illustrated by participants 2 and 3 in Fig. \ref{fig:trial-reread}, who very briefly mouse over ``option B'' before reading the prompt, some mouse movements across a section are too brief to represent genuine reading. Instead, they likely stem from lapsed mouse control.}
Re-reading occurs in roughly half of trials (54\%), revealing three notable patterns: 

\textbf{(1) Annotators most often re-read their chosen response.} They re-read it $38.4\%$ of cases (compared to $25.4\%$ re-read rates for rejected responses and $26.1\%$ for prompts). Relatedly, \textbf{annotators frequently read their chosen response last}:
$74.0\%$ of annotators who re-read any section returned to their chosen response last, suggesting a final confirmation check before submission.

\textbf{(2) Re-reading varies substantially across annotators.} 
Few participants re-read in every trial or in non; most re-read for only a portion of 10 trials, indicating individual differences in review strategies.

\begin{figure}[ht]
    \centering
    \includegraphics[trim={0 0.2cm 0 0},clip,width=0.8\linewidth]{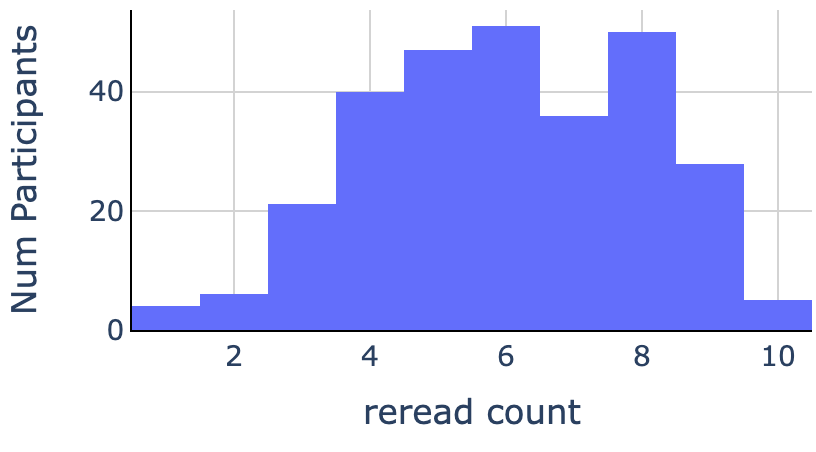}\vspace{-3mm}
    \caption{Each annotator completes ten trials. Across the ten trials, few annotators never (reread count = 0) or always (reread count = 10) reread.
    }
    \label{fig:reread-hist}
\end{figure}

\begin{figure}[ht]
    \centering
    \includegraphics[trim={1cm 1.4cm 0 0},clip,width=0.85\linewidth]{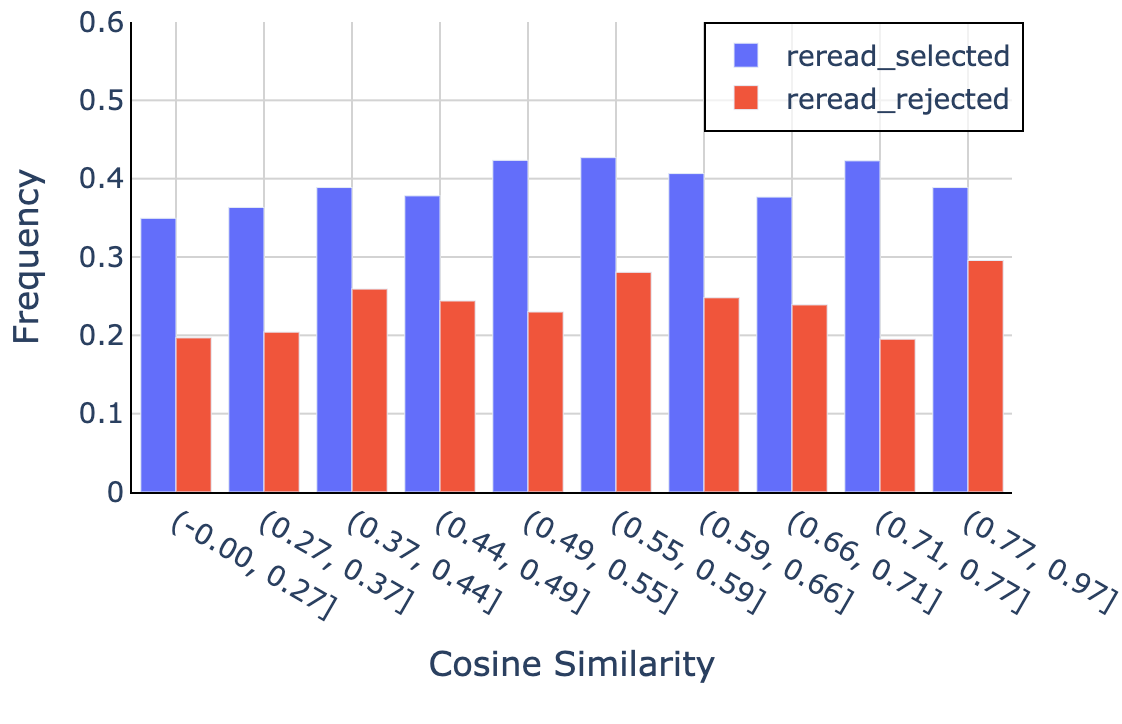}\vspace{-1mm}
    \caption{Frequency of response re-reading behavior ($y$ axis) across quantiles of cosine similarities ($x$ axis) between the two responses. Re-reading frequencies do not correlate with response similarity.
    }
    \label{fig:cosine-sim}
\end{figure}

\textbf{(3) Re-reading does not correlate with response semantic similarities.} Because highly similar responses may lead to more re-reading, we use Sentence-BERT \citep{reimers-2019-sentence-bert} to compute cosine similarity between response pairs and examine re-reading frequencies across cosine similarities. We find that response semantic similarity does not correlate with re-reading frequencies (see Figure~\ref{fig:cosine-sim}). 

\textbf{(4) Reading paths}
For a more detailed view of how annotators navigated the text, we represent each annotator's sequence of visited sections as a \textit{reading path}.
The average path length is $2.47$ ($SD=0.97$), and $17.73\%$ of annotators form a \textit{loop} -- a back-and-forth traversal between the two responses.
Longer paths and loops indicate greater deliberation or indecision between candidates. 



\subsection{Text Skipping Tendencies}\label{sec:skipping}
\begin{figure}
    \centering
    \includegraphics[width=0.9\linewidth]{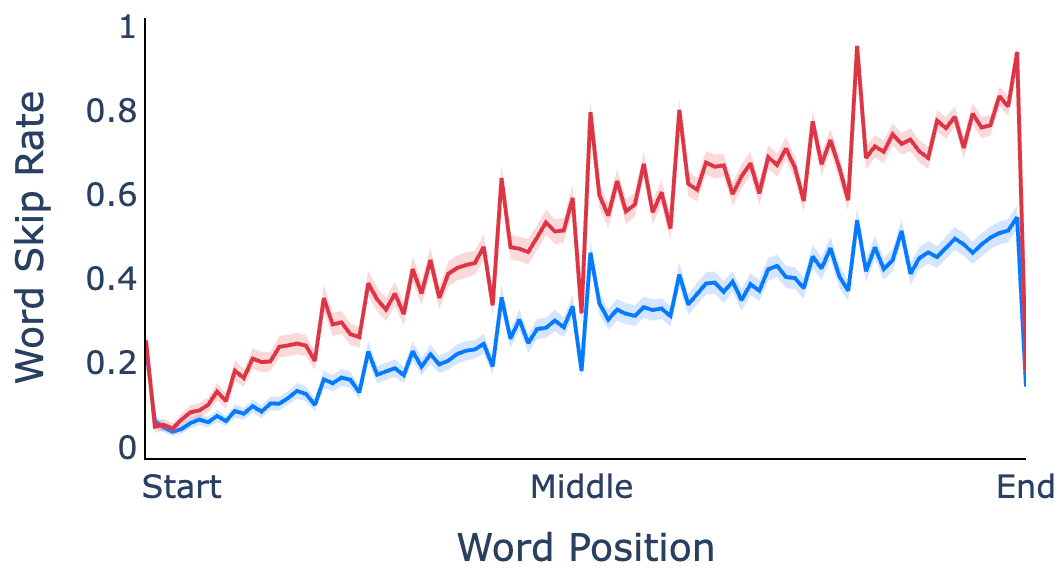}
    \caption{Annotators rarely skip words at the beginning of a response (outside of the initial ``Assistant:'' word), but as the response continues, they more frequently skip words. This is especially pronounced in rejected responses. Word position is calculated via the word index relative to the response length.
    }
    \label{fig:skipping}
\end{figure}

We find that annotators do not always read to the end of the text: \textbf{annotators often make their decision based on an unfinished reading of the responses}, which we term \textit{cognitive economy in reading}.
A paired t-test ($t=-8.52, p < 0.001$) confirms that annotators skip significantly more words in rejected responses ($M = 21.71$) than in chosen one ($M = 15.63$).
We conjecture that annotators often recognize less suitable responses early, leading them read less of the rejected than the chosen response.
We also find that skipped words cluster near the ends of responses (see Figure~\ref{fig:skipping}), suggesting that readers often form judgments \textit{mid-text}.

Annotators also \textbf{read a slightly larger proportion of the \textit{prompt} than of either response}. 
A paired t-test showed significant difference ($p < .01$) between readers' coverage of the prompt relative to either response (response coverage $M = .55$ vs. prompt coverage $M = .59$). 
Recall that we rarely expect coverage to reach 100\% due to the low likelihood that readers will spend a non-trivial amount of time moused over words like `a' or `the,' or highly predictable words like the speaker labels (``Human:'', ``Assistant:'').



\subsection{Agreement and Behavior}\label{sec:agreement}
We assess how reading metrics relate to annotator agreement. First, we discuss overall inter-annotator agreement (\S\ref{sec:iaa}). Second, we examine how annotator agreement relates to both categorical and continuous reading metrics (\S\ref{sec:catmetrics}, \S\ref{sec:contmetrics}). Participant impressions of mouse-tracking are discussed in \S\ref{sec:feedback}.

\subsubsection{Inter-annotator Agreement}\label{sec:iaa}
We measure the overall inter-annotator agreement (IAA) with Krippendorf's alpha \citep{krippendorff2018content}. 
We find $\alpha = 0.25$, which indicates fair agreement among annotators and falls in the range of alphas for subjective annotation tasks reported in \citet{wong2021cross}. 
This is also consistent with recent findings in difficulty of the preference task \cite{cao2025latentcollectivepreferenceoptimization}.

We cannot directly compare IAA with the original dataset since each item had only one annotator.
However, \citet{bai2022training} reported $63\%$ agreement with their annotators, while ours align $59\%$ of the time, suggesting a reasonable level of agreement.

\subsubsection{Categorical metrics}\label{sec:catmetrics}
We applied Chi-square tests of independence to assess whether categorical variables were associated with differences in annotator agreement rates. To determine significance, we set $\alpha = 0.05$ and apply the Bonferroni correction to $p$ values to account for multiple comparisons.

We find that
\textbf{re-reading is significantly more common among annotators who \textit{agree}} ($\chi^2(1) = 11.25, p = 0.001$), suggesting that deliberate review aligns with higher-quality annotations. However, \textbf{looping between two responses is significantly more common among annotators who \textit{disagree}} -- i.e., those who go back and forth at least twice between the two responses have higher rates of disagreement, suggesting that looping reflects high indecision ($\chi^2(1) = 9.42, p = 0.002$). 

Because we also record participants' stated rationale for each annotation (More Helpful, More Accurate, More Concise, Less Harmful, Other), we investigate whether shared rationale is associated with higher agreement. We find the association between rationale choice and agreement approaches significance ($\chi^2(4)=9.25, p=0.055$).

We also assess whether annotators exhibit a bias toward choosing the first or second response and find \textbf{no bias toward choosing the first or second response} ($\chi^2 = 0.29, p = 0.593$).

\subsubsection{Continuous metrics}\label{sec:contmetrics}
To examine the relationship between annotator agreement and continuous metrics (e.g., path length), we form two groups of annotators: pairs who agreed with one another and pairs who did not. We then compare the continuous metric values between these two groups with an independent t-test, again setting $\alpha = 0.05$ and applying the Bonferroni correction to account for multiple comparisons.

Annotators who \textbf{disagree} show  \textbf{slightly longer path lengths} ($M = 2.56$ vs. $M = 2.42$; $t=-3.3, p=0.001$) and  \textbf{slower reading times} ($M = 338.02$ ms/word vs. $M = 315.25$ ms/word; $t = -2.73,\ p = .006$).
Word coverage, on the other hand, does not differ significantly ($M_{agree} = 0.58$ vs. $M_{disagree} = 0.57$; $t = -0.4, p=0.683$), indicating that skimming strategies do not necessarily harm annotation quality.

We also consider the \textit{word focus overlap} between annotators, as annotators who agree may spend more time focusing on the same subset of words. 
For each annotator, we consider the focused words to be those with at least a short duration (for details on how this is computed, see \S\ref{sec:human-attention}). 
Then, we assess the similarity of the focused words between each annotator pair with the Jaccard index (i.e., the intersection over union of the sets of focused words from each annotator). 
In this case, we find that \textbf{the word focus overlap is slightly higher among agreeing annotators} ($M = 0.16 \text{ vs. } M = 0.13$); this difference approaches significance ($t = 2.33, p = 0.021$).

\subsection{Participant experience}\label{sec:feedback}
We conducted a pilot study in which participants tested the interface. 
We were particularly interested in any comments from participants about the mouse tracking, since the paradigm has previously been used in psychology studies that require reading only one to two sentences, and not for longer texts in conjunction with annotation tasks.
In the post-study survey (68 participant responses), 20 participants mentioned \textit{mouse tracking}:
Of these, 8 were positive (e.g., ''fun,'' ''interesting,'' or perception of increased focus due to the small window of revealed text), 7 negative (due to slower reading speeds from mouse tracking), and 5 neutral. Since the primary complaint from participants is that they feel they are slowed down by the mouse tracking, we recommend future adapters adjust monetary compensation for participants accordingly. 

Overall, participants tolerated the mouse tracking paradigm well, even for longer texts, without causing substantial discomfort or annoyance.  This suggests that mouse tracking is feasible for longer, cognitively demanding annotation tasks.

\section{Discussion} 

In \datasetname, we extract several categorical and continuous metrics to augment preference annotations based on the reading processes observed via a mouse tracking interface. 
This yields some important insights regarding how annotators approach the preference reading task in general. 

First, we note a pattern of cognitive economy in reading for annotators, who have a tendency to skip the latter portion of responses. 
This is more pronounced in the case of rejected responses, indicating that annotators do not feel the need to completely finish a response to decide that it is inferior. 
While we don't find that reading less of a response is associated with higher annotator disagreement in our dataset, in some cases, it is possible that this tendency could result in annotators overlooking relevant information. 
Annotation task designers may consider encouraging readers to finish reading both responses, e.g., by including comprehension questions covering the latter part of responses. 

We divide each preference dataset item into three sections: prompt, rejected response, and selected response. We note that annotators re-read sections -- i.e., exit and then later re-enter the same section -- in about half ($54\%$) of all trials. 
Overall, annotators are more likely to re-read their chosen response relative to the rejected response, indicating that annotators may prefer to confirm the quality of their selected response rather than verify the deficiency of their rejected response.

A difficult question in modeling annotator disagreement in subjective tasks like preference selection is whether the disagreement stems from differing opinions, low annotator confidence, or inconsistencies in annotation quality. 
The augmented annotations in \datasetname provide a unique signal in differentiating between potential sources of disagreement.
We find significant associations between re-reading and agreement, as well as significant associations between longer paths and looping behaviors with disagreement.
Such insights can inform annotation post-processing, e.g., annotators who re-read repeatedly can be flagged as potentially low confidence, whereas annotations stemming from an abnormally fast reading process may be flagged as potentially low-quality. 
Using reading patterns in conjunction with demographic data may enhance pluralistic opinion modeling.

Future research can also explore methods to explicitly integrate reading signals, such as word-level scores and reading strategies, into preference model training and reinforcement learning paradigms. For example, re-reading patterns can be explicitly modeled through repetitive encoding of observed behaviors or guided chain-of-thought reasoning to enhance decision-making in preference modeling, and skipped word indices could be leveraged as an auxiliary objective to improve preference decision-making by identifying and filtering out unimportant words.

Finally, while the re-reading behaviors and path definitions discussed here are specific to preference annotations, our methods can be adapted to capture reading behaviors in other challenging and subjective tasks. For instance, domains such as law and medicine involve complex documents with multiple articles that can naturally delineate into sections and high-level reading paths. Such expert annotation tasks require careful engagement from skilled annotators, making them well-suited to the reading augmented annotations we propose here.

\section{Conclusion}

We introduce \datasetname, a first-of-its-kind dataset that contains the reading process of annotators as they complete the preference annotation task. Reading processes are captured via a mouse-tracking-enabled annotation interface, demonstrating the feasibility of crowd-sourced, large-scale reading behavior collection for NLP annotations. 
\datasetname offers several insights into annotator reading behaviors during the preference annotation task, and we find several relationships between reading processes and annotator agreement. Together, our results suggest that the proposed augmentation strategy provides a novel dimension in understanding human thought processes during an annotation task.

\section*{Limitations}
The mouse-tracking interface is less naturalistic than typical reading, and it may impose additional task demands on annotators. 
However, we consider this a worthwhile trade-off, as it would otherwise be impractical to gather large-scale data on annotators’ reading processes. 
Accordingly, our findings should be interpreted within the context of this specific experimental paradigm.

\section*{Ethics Statement}
Mouse-tracking data, like all behavioral data, carries a potential risk of revealing participant identities. All data in this study were collected with informed consent, anonymized prior to analysis, and handled in accordance with institutional review protocols. No personally identifiable information was retained or shared.

\section{Bibliographical References}\label{sec:reference}

\bibliographystyle{lrec2026-natbib}
\bibliography{lrec2026-example}

\section{Language Resource References}
\label{lr:ref}
\bibliographystylelanguageresource{lrec2026-natbib}
\bibliographylanguageresource{languageresource}

\end{document}